\documentclass{article}
\usepackage{ijcai20}

\usepackage{times}
\usepackage{soul}
\usepackage{url}
\usepackage[hidelinks]{hyperref}
\usepackage[utf8]{inputenc}
\usepackage[small]{caption}
\usepackage{graphicx}
\usepackage{amsmath}
\usepackage{amsthm}
\usepackage{booktabs}
\usepackage[ruled,vlined,linesnumbered]{algorithm2e}

\usepackage{amsthm,amsmath,amsfonts,bm,xspace}
\usepackage{upgreek}
\usepackage{color}

\def\vs{{\em v.s.}\xspace}
\newcommand{\comment}[1]{}

\def\eqref#1{(\ref{#1})}

\def\1{\bm{1}}

\def\vtheta{{\bm{\theta}}}
\def\vpsi{{\bm{\psi}}}
\def\vphi{{\bm{\phi}}}

\def\vF{{\bm{F}}}

\def\vq{{\bm{q}}}

\def\vs{{\bm{s}}}

\DeclareMathAlphabet{\mathsfit}{\encodingdefault}{\sfdefault}{m}{sl}
\SetMathAlphabet{\mathsfit}{bold}{\encodingdefault}{\sfdefault}{bx}{n}

\title{Retrieve, Program, Repeat: Complex Knowledge Base Question Answering via Alternate Meta-learning}

\author{
Yuncheng Hua$^{1,3}$
\and
Yuan-Fang Li$^2$\and
Gholamreza Haffari$^2$\and
Guilin Qi$^{1,4}$\footnote{Contact Author}\And
Wei Wu$^1$\\
\affiliations
$^1$School of Computer Science and Engineering, Southeast University, Nanjing, China\\
$^2$Faculty of Information Technology, Monash University, Melbourne, Australia\\
$^3$Southeast University-Monash University Joint Research Institute, Suzhou, China\\
$^4$Key Laboratory of Computer Network and Information Integration, Southeast University, China
\emails
\{devinhua, gqi, wuwei\}@seu.edu.cn,
\{yuanfang.li, gholamreza.haffari\}@monash.edu
}

\begin{document}

\maketitle

\begin{abstract}
%
A compelling approach to complex question answering is to convert the question to a sequence of actions, which can then be executed on the knowledge base to yield the answer, aka the programmer-interpreter approach. 
Use similar training questions to the test question, meta-learning enables the programmer to adapt to unseen questions to tackle potential distributional biases quickly.
However, this comes at the cost of manually labeling similar questions to learn a retrieval model, which is tedious and expensive.
In this paper, we present a novel method that automatically learns a retrieval model alternately with the programmer from weak supervision, i.e., the system’s performance with respect to the produced answers.
To the best of our knowledge, this is the first attempt to train the retrieval model with the programmer jointly.
Our system leads to state-of-the-art performance on a large-scale task for complex question answering over knowledge bases.
We have released our code at {\small\textsf{\url{https://github.com/DevinJake/MARL}}}.
\end{abstract}

\section{Introduction}
\label{Intro}
%
Complex question answering (CQA) over knowledge base (KB) aims to map natural-language questions to logical forms (\emph{annotations}), i.e., programs or action sequences, which can be directly executed on the KB to yield answers (\emph{denotations})~\cite{berant2013semantic,shen-etal-2019-multi}.
%
Different from other forms of KBQA, such as multi-hop question answering, CQA requires discrete \emph{aggregation} actions, such as set intersection/union, counting, and min/max, to yield answers, which can be entities in the KB as well as numbers. 

\begin{figure}[htb]
\centering
\includegraphics[width=0.46\textwidth]{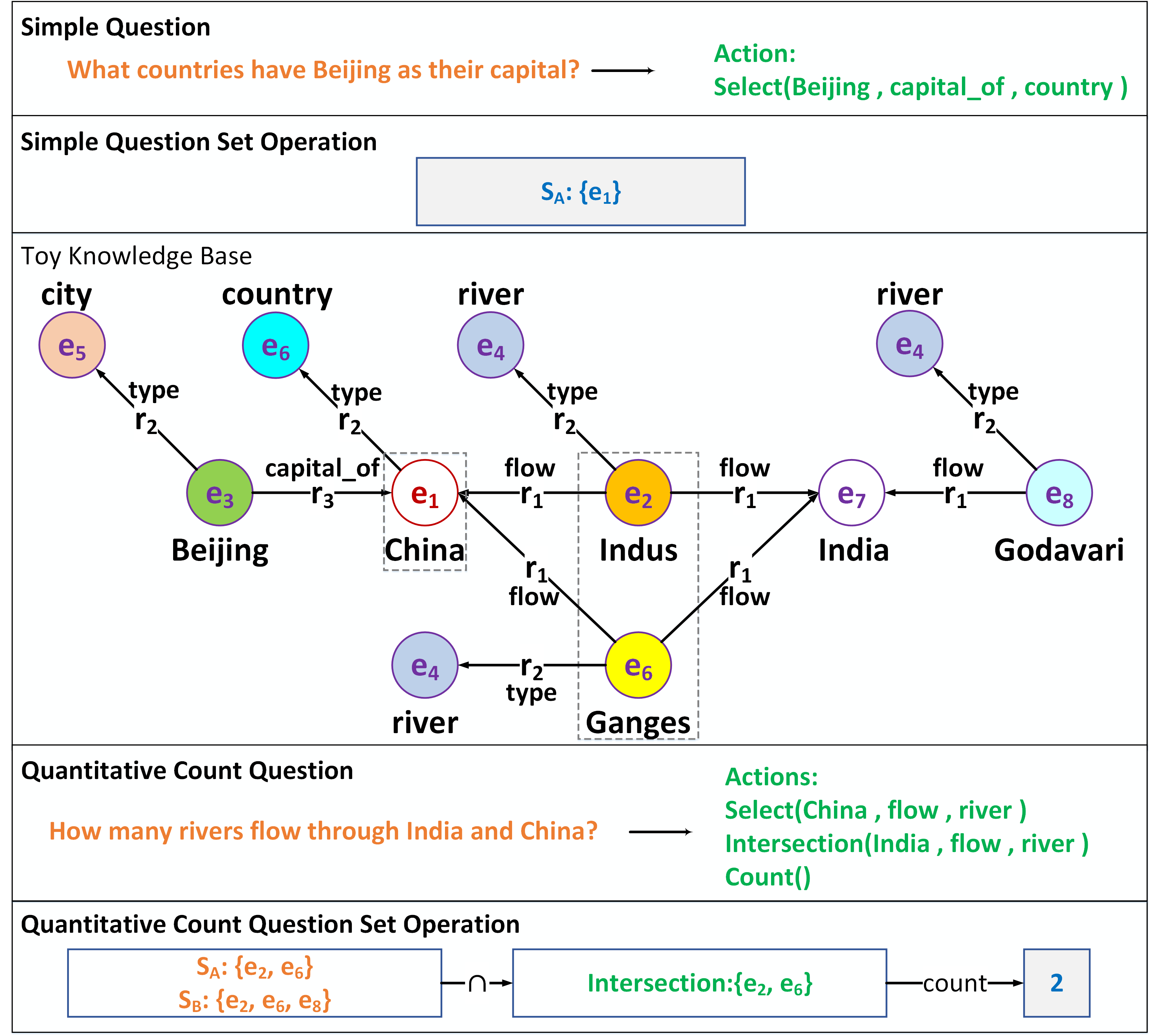}
\caption{Two questions of different types in the CQA dataset.} \label{fig:example}
\end{figure}

Taking one dataset CQA~\cite{saha2018complex} as an example, the questions are organized into seven categories, and the questions in the different categories vary substantially in length and complexity.
For instance, as shown in Figure~\ref{fig:example}, the question in the `Simple' category only requires one `select' action to answer, while the question in the `Quantitative Count' category requires the execution of the sequence of actions `select', `intersection' and `count' to obtain the answer.
%
%

Standard KBQA approaches~\cite{berant2013semantic,yih2014semantic,bordes2015large,yu2017improved,guo2018dialog,saha2018complex,ansari2019neural} adopt imitation learning, memory network, or RL, and they typically train one model that fits the entire training dataset and use it to answer all questions.
Such a one-size-fits-all model aims to learn the generic knowledge across the questions in the training phase and use the knowledge to predict action sequences for each test question in the inference phase.
However, when the training questions vary widely, less common knowledge will be shared across them. Thus the increase in the diversity of the questions is associated with a reduction in the generic knowledge. 
It would be challenging for the one-size-fits-all model to produce optimal logical forms for each instance. 

Several methods have recently been proposed to address this challenge. 
Complex Imperative Program Induction from Terminal Rewards (CIPITR)~\cite{saha2019complex} takes a neural program induction (NPI) approach and proposes to train different independent models, one for each category of CQA questions.
Essentially, CIPITR aims to learn a one-size-fits-all model for each \emph{question type} instead of granting the model the ability to transfer from one question type to another. 

S2A~\cite{DBLP:conf/acl/GuoTDZY19} attempts to address this challenge with a method based on a meta-learning method~\cite{urgen1992learning}, employing Model Agnostic Meta-Learning (MAML)~\cite{finn2017model} specifically. 
S2A puts forward a retriever and a meta-learner. 
It first collects annotations for each question in the training set and trains a retriever to find questions with similar annotations.
Subsequently, in training the meta-learner, when faced with a new question, the already-trained retriever finds samples from the training set that are similar to the new question and regards these samples as a support set.
The meta-learner views the new question along with the support set as a new task and then learns a specific model adaptive to the task.
%

However, S2A is faced with two difficulties.
Firstly, as it employs a teacher forcing approach in the process of training the retriever, it places a burden on collecting annotations for each question. 
Secondly, the retriever is trained independently of the meta-learner. 
Thus, the result of the meta-learner answering the questions is irrelevant to the retriever.
Therefore it is hard to evaluate the quality of the support set that the retriever establishes for each new question and consequently tricky to measure the impact of the retriever on answering the questions. 
If the samples found by the retriever are not similar to the new question as expected, the meta-learner will be misguided by the deviated examples and thus learns a model that is not suitable for the current task.

Furthermore, though the approach taken by S2A is the most similar to ours, the tasks differ significantly. 
S2A aims to answer \emph{context-dependent} questions, where each question is part of a multiple-round conversation. 
Hence, the context-aware retriever proposed in S2A considers the relevant conversation.
On the contrary, however, in this paper, we consider the setup where the questions are single-round and directly answerable from the KB.
Thus, a novel challenge arises in retrieving accurate support sets without conversation-based context information.
In this work, to address the above problems, we propose MetA Retrieval Learning (MARL), a new learning algorithm that jointly optimizes retriever and programmer in two stages. 

In the first stage, we fix the parameter of the retriever and employ it to select the top-$N$ similar questions (secondary questions) to a given target question (primary question).
The \emph{trial} trajectories, along with the corresponding rewards for answering each secondary question, are used to adapt the programmer to the current primary question.
The feedback on how correctly the adaptive programmer answers the primary question is used to update the weights of the programmer.

In the second stage, we fix the parameter of the programmer and optimize the retriever. 
The \emph{general} programmer first outputs an answer and gain a reward to the primary question without using any secondary questions.
Then we sample $M$ different sets of secondary questions following the retriever policy and employ the question sets to learn $M$ different programmers.
Each specific programmer will generate an answer to the primary question and gain some reward.
We consider the difference between the reward yielded by the general programmer and the reward gained by each adapted programmer as the contribution of employing the corresponding support set. 
Thus, the reward difference provides the training signal to optimize the policy of the retriever: positive difference increases the probability that a set of secondary questions is chosen, and a negative difference lowers the probability.
%
%

We train the programmer and the retriever alternatively until the models converge.
Note that in our method, the training of the retriever is done in a weakly-supervised manner.
The retriever is optimized to find better support sets according to the programmer's performance of answering primary questions, rather than employing teacher forcing. 
Since one support set generates one adaptive programmer, we employ the feedback obtained by evaluating the adaptive programmer as a weak supervision signal to optimize the retriever.
Therefore we encourage the retriever to find the support set that leads to a superior programmer that gains more reward.
At the same time, the programmer is optimized alongside the retriever.

%
%
We evaluate our method on CQA~\cite{saha2018complex}, a large-scale complex question answering dataset. 
MARL outperforms standard imitation learning or RL-based methods and achieves new state-of-the-art results with overall micro and macro F1 score. 
Notably, MARL uses only 1\% of the training data to form the tasks for meta-learning and achieve competitive results.

\section{Alternate Meta-learning for Complex Question Answering over Knowledge Bases}
\label{approach}
We now introduce our method for combining weakly-supervised {\it retriever} and {\it meta-learning} to solve the CQA task, which we call MARL. 

In this paper, we consider each new complex question in the training dataset as one \emph{primary question} and view answering one primary question as an individual task. 
To solve an unseen task effectively, we aim to build a unique {\it programmer} for each task. 
Thus we harness the {\it retriever} to find the top-N most similar questions as the \emph{secondary questions}, and propose a meta reinforcement learning approach to rapidly adapt the programmer to the primary question with the support of the secondary questions.

\subsection{Method Overview}
The goal of MARL is to train the retriever to find the optimal secondary questions for the programmer, which, when faced with a new primary question, can quickly adapt the programmer to the unseen question and effectively improve QA performance. 
To accomplish this goal, we train two networks jointly: (1) an encoder-decoder network, which is viewed as the {\it programmer}, that transforms questions into programs; and (2) a {\it retriever} network, which finds the secondary questions, i.e., the N questions that are the most analogous to the primary question. 

\begin{figure}[htb]
\centering
\includegraphics[width=0.5\textwidth]{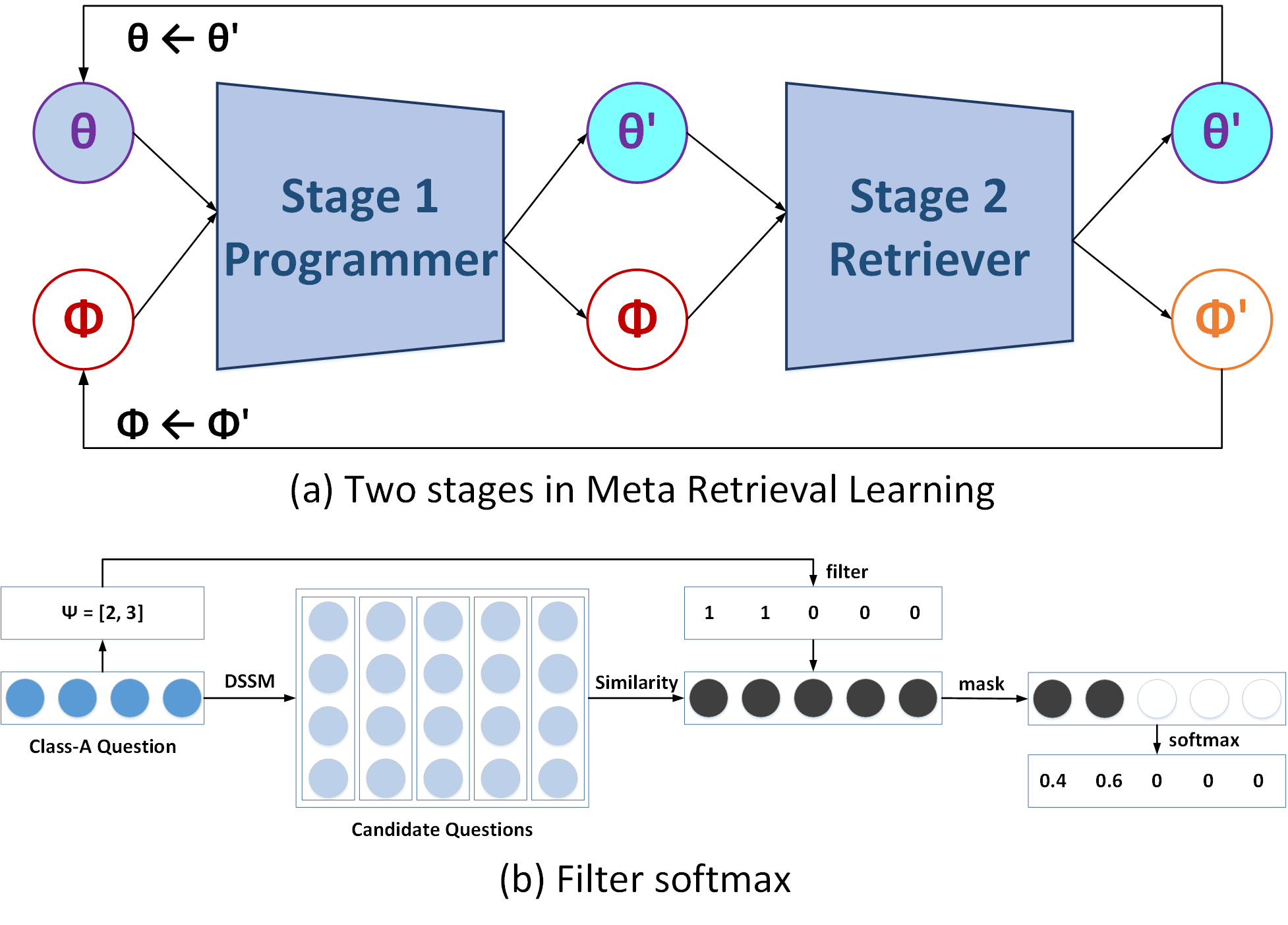}
\caption{(a). Illustration of the two stages in the training process of MARL. (b) Illustration of the filter softmax.} \label{fig:framework}
\end{figure}

We denote the encoder-decoder network as a policy $\pi(\boldsymbol{\tau}|{\vq_i;\vtheta})$ with parameter $\vtheta$, and the retriever network as a policy $\pi(\vq_{c_i}|\vq_{pri};{\vphi})$ with parameter $\vphi$. 
Both networks take questions as the input, while the output of the programmer network is the programs that can be directly executed on the KB to generate answers, and the production of the retriever network is the probability distribution over all candidate training samples. 
For the retriever network, the N samples that have the highest probabilities are selected as secondary questions for a given primary question. 
Both the parameters $\vtheta$ and $\vphi$ are optimized by the rewards, which are the comparison between the generated answers and the ground-truth answers.

In the training process, when the primary question and the corresponding secondary questions are similar (i.e., with analogical structures and semantic meanings), the programmer will be guided more effectively in finding the optimal direction of parameter update.
On the contrary, if the secondary questions are randomly extracted from the training set, they might share a little commonality with the primary question and consequently yield disparate action sequences. 
Since only a small amount of samples are used to finetune the adaptive model, such dissimilar instances will lead the programmer on a noisier (more random) path for gradient update.

In the retriever network, we thus cluster the training samples into several groups based on the question type and use a vector $\vpsi$ to record group membership of each question.
For each primary question, based on $\vpsi$, a unique {\it filter} $\vF$ will be generated to determine the possible questions as candidate secondary questions by filtering out irrelevant questions. 
Given the primary question $\vq_{pri}$, the retriever network computes the probability of choosing a candidate question $\vq_{c_i}$ as a secondary question by applying a filter softmax function, which could be denoted as
\begin{equation}
\pi(\vq_{c_i}|\vq_{pri};{\vpsi},{\vphi})
\label{softmax}
\end{equation}

Following the probability, we sample a set of secondary questions and denote them as $s^{\vq_{pri}}$, which is used to support adapting the programmer for the primary question $\vq_{pri}$.

\subsection{Model Objectives}
We consider two types of knowledge that can contribute to the answering of a new question: the task-specific knowledge, which is the particular features shared with a small number of similar questions that inform what the current task is; and the task-agnostic knowledge, which is the generic features shared across different tasks. 
If similar questions are accurately selected, the model could effectively exploit the particular features among these questions and acquire task-specific knowledge to recognize the new task. 
If the generic features are suitable for many tasks, the model could rapidly adapt itself to a new task and produce satisfactory results for the task by conducting only a small number of gradient updates. 
We employ a retriever to find similar questions for acquiring the task-specific knowledge and train a generic policy to accumulate task-agnostic knowledge.

In our method, the performance of answering a new question is represented by the reward feedback on whether the question is correctly answered. 
It is hard to disentangle the contribution of the two types of knowledge. 
In other words, it is difficult to determine which factor is the reason for generating a correct answer, the task-specific knowledge gained from similar questions, or the task-agnostic knowledge acquired from the well-learned generic policy. 

As shown in Figure~\ref{fig:framework}(a), we split the training process in each epoch into two independent stages and train the programmer and the retriever alternately. 
In the {\bf first stage}, we fix the parameter $\vphi$ of the retriever network and employ it to select secondary questions to optimize the parameter $\vtheta$ of the programmer. 
The target of the first stage is to encourage the model to learn the generic knowledge that is broadly applicable to all tasks by updating $\vtheta$. 
In the {\bf second stage}, we fix the parameter $\vtheta$ of the programmer and update the retriever network by computing its gradients with respect to the programmer's output. 
We compare the results of answering the same primary question with or without using the retriever and consider the difference between the results as the indication of how the retriever contributes to solving the primary question. 
Such differences are used to train the retriever to find the optimal set of secondary questions (the \emph{support set}). 
Both networks are trained iteratively until convergence.

The training approach is depicted in Algorithm \ref{MARL}.

In the \textbf{first stage} (lines~\ref{alg:stg1_start}--\ref{alg:stg1_end} in Algorithm~\ref{MARL}) of each epoch, we propose a {\it meta reinforcement learning} (meta-RL) approach to update $\vtheta$, i.e.\ the programmer. 
We use the gradient-based meta-learning method to solve the meta-RL problem such that we can obtain the optimal policy for a given task after performing a few steps of vanilla policy gradient (VPG) with the task.
We employ \emph{Monte Carlo integration} (MC) as the approximation strategy in the Policy Gradient.      

The meta-learning process is divided into two steps to solve a task, namely the meta-training step and the meta-test step. 
Suppose we are trying to answer the primary question $\vq_{pri}$, $N$ secondary questions $s^{\vq_{pri}}$ will be first found by the retriever network, and we consider $\vq_{pri}$ together with $s^{\vq_{pri}}$ as a pseudo-task $\mathcal{T}_{pse}$.
During meta-training, the meta-RL model first generates $K$ trajectories for each question in $s^{\vq_{pri}}$ based on parameter $\vtheta$. 
The reward of each trajectory is given by the environment and subsequently used to adapt $\vtheta'$  to task $\mathcal{T}_{pse}$ as follows:
\begin{equation}
\vtheta' \leftarrow \vtheta+\eta_1 \nabla_{\vtheta} \sum_{\vq_i \in \vs^{\vq_{pri}}}
\mathbb{E}_{\boldsymbol{\tau} \sim \pi(\boldsymbol{\tau}|{\vq_i;\vtheta})}[R(\boldsymbol{\tau})]
\label{adapted_theta}
\end{equation}

In the meta-test step, another $K$ trajectories corresponding to the primary question $\vq_{pri}$ are further produced by $\vtheta'$. 
The reward produced by the new trajectories is considered as the evaluation of the adapted policy $\vtheta'$ for the given task $\mathcal{T}_{pse}$; thus, we have the following objective: 
\begin{equation}
J_{\vq_{pri}}(\vtheta')\overset{\textrm{def}}{=} \mathbb{E}_{\boldsymbol{\tau}' \sim \pi({\boldsymbol{\tau}}'|\vq_{pri};\vtheta')}[R({\boldsymbol{\tau}}')]
\label{meta_test_loss}
\end{equation}

The parameter of the generic policy $\vtheta$ are then trained by maximizing the objective $J_{\vq_{pri}}(\vtheta')$, 
\begin{equation}
\vtheta\leftarrow\vtheta + \eta_2 \nabla_{\vtheta}J_{\vq_{pri}}(\vtheta')
\label{updated_theta}
\end{equation}

In each VPG step, since we have $N$ samples in $\vs^{\vq_{pri}}$, we use $N$ policy gradient adaptation steps to update $\vtheta'$.
Meanwhile, we use one policy gradient step to optimize $\vtheta$ based on the evaluation of $\vtheta'$.
We denote the optimized parameter of the programmer $\vtheta^*$.

In the {\bf second stage} (lines~\ref{alg:stg2_start}--\ref{alg:stg2_end} in Algorithm~\ref{MARL}) of each epoch, we propose a {\it reinforcement learning} approach to update $\vphi$, i.e.\ the retriever.
The primary questions that we want to solve are the same as the data used in the first stage. 
When answering the primary question $\vq_{pri}$, we first generate a trajectory based on parameter $\vtheta^*$ that has been optimized in the first stage as:
\begin{equation}
\boldsymbol{\tau}^* \leftarrow \textrm{decode}(\pi(\boldsymbol{\tau}|\vq_{pri};{\vtheta^*}))
\label{greedy_decode}
\end{equation}
and further, obtain the reward $R(\boldsymbol{\tau}^*)$ by executing the trajectory.

We then employ the retriever to compute the probability of selecting one candidate question as a secondary question.
From this distribution, we sample $N$ secondary questions to form a support set, rather than directly choosing the $N$ questions with the highest probability.
The probability of sampling a set of questions $\vs^{\vq_{pri}}$ is:
\begin{equation}
P(\vs^{\vq_{pri}})=\prod_{\vq_{c_i} \in \vs^{\vq_{pri}}} \pi(\vq_{c_i}|\vq_{pri},{\vpsi};{\vphi})
\label{sample_supportset_probability}
\end{equation}

We could repeat sampling several times and acquire different support sets.
Employ one set $\vs^{\vq_{pri}}_m$ from these support sets, as what we have done in the first stage, we get the adapted $\vtheta^{\ast'}_m$:
\begin{equation}
\vtheta^{\ast'}_m \leftarrow \vtheta^*+\eta_1 \nabla_{\vtheta^*} \sum_{\vq_i \in \vs^{\vq_{pri}}_m} \mathbb{E}_{\boldsymbol{\tau} \sim \pi(\boldsymbol{\tau}|{\vq_i;\vtheta^*})}[R(\boldsymbol{\tau})]
\label{adapted_theta_for_retriever}
\end{equation}

Then we generate a new trajectory under $\vtheta^{\ast'}_m$ for the primary question:
\begin{equation}
\boldsymbol{\tau}^{\ast'}_m \leftarrow \textrm{decode}(\pi(\boldsymbol{\tau}|\vq_{pri};{\vtheta^{\ast'}_m}))
\label{greedy_decode_retriever}
\end{equation}
and also compute the reward for this trajectory as $R(\boldsymbol{\tau}^{\ast'}_m)$.

We regard the difference between $R(\boldsymbol{\tau}^*)$ and $R(\boldsymbol{\tau}^{\ast'}_m)$ as the contribution of the support set $\vs^{\vq_{pri}}_m$, which is used to learn the task-specific knowledge from the questions in $\vs^{\vq_{pri}}_m$.

\begin{algorithm}[htb]
\caption{The MARL algorithm}
\label{MARL}
\DontPrintSemicolon
\KwIn{Training dataset $Q_{train}$, step size $\eta_1$, $\eta_2$, $\eta_3$}    
\KwOut{The learned $\vtheta^*$ and $\vphi^*$}
Initialize groups vector $\vpsi$\;
Randomly initialize $\vtheta$\ and $\vphi$\;
\While{not converged}{
    \ForEach{training iteration}
        {\label{alg:stg1_start}
        Sample a batch of primary questions $Q_{pri} \sim Q_{train}$\;
        \ForEach{$\vq_{pri} \in Q_{pri}$}
        {
        Retrieve $\vs^{\vq_{pri}}$ with $\vphi$ and $\vpsi$\;
        $\mathcal{L} = \sum_{\vq_i \in \vs^{\vq_{pri}}}
\mathbb{E}_{\boldsymbol{\tau} \sim \pi(\boldsymbol{\tau}|{\vq_i;\vtheta})}[R(\boldsymbol{\tau})]$\;
        Get adapted parameters: $\vtheta'\leftarrow \vtheta+\eta_1 \nabla_{\vtheta}\mathcal{L}$\; 
        $J_{\vq_{pri}}(\vtheta')\overset{\textrm{def}}{=} \mathbb{E}_{\boldsymbol{\tau}' \sim \pi({\boldsymbol{\tau}}'|\vq_{pri};\vtheta')}[R({\boldsymbol{\tau}}')]$\;
        }
        Update $\vtheta\leftarrow\vtheta+\eta_2\nabla_{\vtheta}\sum_{\vq_{pri} \in Q_{pri}} J_{\vq_{pri}}(\vtheta')$\;
        }
    $\vtheta^* \leftarrow \vtheta$\label{alg:stg1_end}\;
    \ForEach{training iteration}
    {\label{alg:stg2_start}
    Sample a batch of primary questions $Q_{pri} \sim Q_{train}$\;
    \ForEach{$\vq_{pri} \in Q_{pri}$}
    {$\boldsymbol{\tau}^* \leftarrow \textrm{decode}(\pi(\boldsymbol{\tau}|\vq_{pri};{\vtheta^*}))$\; 
    Compute reward $R(\boldsymbol{\tau}^*)$\;
    Sample support sets $\mathcal{M}$ with $\vphi$ and $\vpsi$\;
    \ForEach{$\vs^{\vq_{pri}}_m \in \mathcal{M}$}
    {
    $\vtheta^{\ast'}_m\leftarrow\vtheta^*+\eta_1\nabla_{\vtheta^*}\sum_{\vq_i\in\vs^{\vq_{pri}}_m}\mathbb{E}_{\boldsymbol{\tau}\sim\pi(\boldsymbol{\tau}|{\vq_i};\vtheta^*)}R(\boldsymbol{\tau})$\;
    $\boldsymbol{\tau}^{\ast'}_m \leftarrow \textrm{decode}(\pi(\boldsymbol{\tau}|\vq_{pri};{\vtheta^{\ast'}_m}))$\;
    Compute reward $R(\boldsymbol{\tau}^{\ast'}_m)$\;
    }
    $J_{\vq_{pri}}(\vphi)\overset{\textrm{def}}{=} \mathbb{E}_{\vs^{\vq_{pri}}_m \sim P(\vs^{\vq_{pri}})}[R(\boldsymbol{\tau}^{\ast'}_m)-R(\boldsymbol{\tau}^*)]$\;
    }
    Update $\vphi\leftarrow\vphi + \eta_3 \nabla_{\vphi}\sum_{\vq_{pri} \in Q_{pri}} J_{\vq_{pri}}(\vphi)$\;
    }
    $\vphi^* \leftarrow \vphi$\label{alg:stg2_end}\;
}
\textbf{Return}  The learned $\vtheta^*$ and $\vphi^*$ 
\end{algorithm}

The retriever network is then updated by encouraging the particular support sets to be chosen such that, if the policy $\vtheta$ is adapted to the current task by using these support sets, the reward of answering the primary question would be maximized.
Therefore, we harness the difference as the reward and have the objective:
\begin{equation}
J_{\vq_{pri}}(\vphi)\overset{\textrm{def}}{=} \mathbb{E}_{\vs^{\vq_{pri}}_m \sim P(\vs^{\vq_{pri}})}[R(\boldsymbol{\tau}^{\ast'}_m)-R(\boldsymbol{\tau}^*)] 
\label{objective_phi}
\end{equation}

The parameter of the retriever network $\vphi$ are then updated by maximizing the objective $J(\vphi)$ as:
\begin{equation}
\vphi\leftarrow\vphi + \eta_3 \nabla_{\vphi}J(\vphi)
\label{update_phi}
\end{equation}

However, it is often infeasible to compute the gradient in Equation~(\ref{objective_phi}) because it involves taking an expectation over all possible sampled support sets. 
Hence, in practice, we employ Monte Carlo integration to approximate the expectation, which is:
\begin{equation}
\Delta_{MC}=\frac{1}{M}\sum_{\vs^{\vq_{pri}}_m \in \mathcal{M}} [R(\boldsymbol{\tau}^{\ast'}_m)-R(\boldsymbol{\tau}^*)-\mathcal{C}]\nabla log(P(\vs^{\vq_{pri}}_m)) 
\label{monte_carlo_c}
\end{equation}
where $\mathcal{M}$ is a collection of $M$ sets of secondary questions retrieved to support the primary question $\vq_{pri}$, and $\mathcal{C}$ is a baseline with a constant value to reduce the variance of the estimate without altering its expectation.

\subsection{Filter Softmax}
As mentioned before, we categorize the instances in the training set based on the question type and employ a vector $\vpsi$ to record the type information of the questions. 
For each question type, based on $\vpsi$, a filter $\vF$ will be created to make the retriever only consider questions of the same type when searching for secondary questions for a primary question of that type. 
The filter value is set to zero for all instances that do not have the same type of a given question~\cite{liu2019self}. 
For example, as shown in Figure~\ref{fig:framework}(b), assume we have a dataset with two question types $y = [A, B]$, and we arrange the dataset to group the same type questions together.
Thus we have a $\vpsi = [2,3]$ to indicate that two questions belong to type $A$ and three questions from type $B$.
In this case, the filter is set as $\vF = [1,1,0,0,0]$ for the primary question of type $A$ and $\vF = [0,0,1,1,1]$ for type $B$.
Applying the filter to the softmax function would mask the irrelevant questions, which have types different from the primary question.
This allows the retriever to reduce the search space and increase the probability of finding expected questions. 
The filter softmax function produces the following probability:
\begin{equation}
p(q_{c_i}|q_{pri}) = \frac{e^{sim(q_{pri},q_{c_i})} \odot \vF_i}{\sum_{i'} (e^{sim(q_{pri},q_{c_i'})}\odot \vF_{i'})}, 
\end{equation}
where $p(q_{c_i})$ represents the probability that the candidate question $q_{c_i}$ is selected as one of the secondary questions for the primary question $q_{pri}$, and $sim(q_{pri},q_{c_i})$ is the semantic similarity between them.
Besides, $\vF_i$ is the binary value in the filter $\vF$ that recognizes whether question $q_{c_i}$ has the same type as $q_{pri}$ or not, and $\odot$ represents element-wise multiplication. 
In our work, Deep Structured Semantic Model (DSSM) method~\cite{huang2013learning} is employed to compute the similarity between two questions. 

\section{Evaluation}
\label{evaluation}
We evaluated our MARL model on the CQA dataset~\cite{saha2018complex}. 
CQA is a large-scale complex question answering dataset containing 944K/100K/156K question-answer pairs for training/validation/test. 
CQA divides itself into seven categories based on the nature of answers, e.g., entities as answers in the `Simple Question' category and numbers in the category `Quantitative (Count)'.
We used `accuracy' as the evaluation metric for questions whose type is `Verification', `Quantitative (Count)', and `Comparative (Count)'; and `F1 measure' to other kinds of questions. 
However, to simplify the presentation and stay consistent with literature~\cite{saha2019complex,ansari2019neural}, we denote `accuracy' as `F1 measure' in Table~\ref{tab:comparison}. 
Hence, the model performance is evaluated on the F1 measure in this paper.
Furthermore, we compute the micro F1 and macro F1 scores for all the models based on the F1-scores of the seven question types.

Also, in our analysis of the CQA dataset, we found that the seven types of questions vary substantially in complexity.
We discovered that `Simple' is the simplest that only requires two actions to answer a question, whereas `Logical Reasoning' is more difficult that requires three actions.
Categories `Verification' and `Quantitative Reasoning' are the next in the order of difficulty, which need 3--4 actions to answer.
The most difficult categories are `Comparative Reasoning', `Quantitative (Count)', and `Comparative (Count)', needing 4--5 actions to yield an answer.
Saha et al.~\shortcite{saha2019complex} drew a similar conclusion in the manual inspection of these seven question categories.

\begin{table*}[htb]
\centering
\begin{tabular}{p{4.3cm}*{7}{r}} 
\toprule
{\textbf{Question category}} & {\textbf{KVmem}} & {\textbf{CIPITR-All}} & {\textbf{CIPITR-Sep}} & \quad\textbf{Vanilla} & {\textbf{Random}}     & \textbf{Jaccard}      & \textbf{MARL}\\ \midrule
Simple Question (462K)       & 41.40\%          & 41.62\%               & \textbf{94.89\%}      & 84.67\%               & 85.22\%               & 86.07\%               & \underline{88.06\%}\\
Logical Reasoning (93K)      & 37.56\%          & 21.31\%               & \textbf{85.33\%}      & 76.58\%               & 78.87\%               & 78.89\%               & \underline{79.43\%}\\
Quantitative Reasoning (99K) & 0.89\%           & 5.65\%                & 33.27\%               & 47.20\%               & 48.11\%               & \underline{48.21\%}   & \textbf{49.93\%}\\
Verification (Boolean) (43K) & 27.28\%          & 30.86\%               & 61.39\%               & 81.94\%               & 85.04\%               & \underline{85.24\%}   & \textbf{85.83\%}\\ \midrule
Comparative Reasoning (41K)  & 1.63\%           & 1.67\%                & 9.60\%                & 58.05\%               & 61.96\%               & \underline{63.07\%}   & \textbf{64.10\%}\\
Quantitative (Count) (122K)  & 17.80\%          & 37.23\%               & 48.40\%               & 60.36\%               & 60.04\%               & \underline{60.47\%}   & \textbf{60.89\%}\\
Comparative (Count) (42K)    & 9.60\%           & 0.36\%                & 0.99\%                & 39.25\%               & 38.50\%               & \underline{39.50\%}   & \textbf{40.50\%}\\ \midrule
Overall macro F1             & 19.45\%          & 19.82\%               & 47.70\%               & 64.01\%               & 65.39\%               & \underline{65.92\%}   & \textbf{66.96\%}\\
Overall micro F1             & 31.18\%          & 31.52\%               & 73.31\%               & 74.72\%               & 75.69\%               & \underline{76.31\%}   & \textbf{77.71\%}\\
\bottomrule
\end{tabular}
\centering
\caption{Performance comparison (measured in F1) on the CQA test set. For each category, best result is \textbf{bolded} and second-best result is \underline{underlined}. The number of instances in each category in the training set is also given next to the category name.}\label{tab:comparison}
\end{table*}

\begin{table}[htb]
\begin{tabular}{lr}
\toprule
{\textbf{Feature}} & {\textbf{Overall micro F1}}\\ 
\midrule
Vanilla & 74.72\% \\
\midrule
MARL (random retriever)  & +0.97\% \\
MARL (Jaccard retriever) & +1.59\% \\ 
MARL (full model)        & +2.99\% \\
\bottomrule
\end{tabular}
\centering
\caption{Ablation study on the test set on macro F1 score change with the addition of meta-learning and different retrievers. Full model (MARL) has micro F1 of 77.71\% as shown in Table~\ref{tab:comparison}.}\label{tab3}
\end{table}

\subsection{Implementation Details}
In the CQA dataset, we randomly sampled approximately 1\% of the training set (10,353 out of 944K training samples) and annotated them with pseudo-gold action sequences with a breadth-first-search (BFS) algorithm~\cite{guo2018dialog}.
We denote this dataset as $Q_{pre}$.
We trained a BiLSTM-based programmer with $Q_{pre}$, and further optimized it through RL with another 1\% unannotated questions from the training set.
We note this RL-based model is a one-size-fits-all model and denote it as \textbf{Vanilla}. 
We randomly selected another 2,072 samples from the 944K training questions to establish pseudo-tasks for meta-learning, which represented only approx.\ 0.2\% of the training set.
This model that jointly learns the programmer along with the retriever by employing MAML is our full model and is denoted \textbf{MARL}. 

We implemented the MARL model in PyTorch with all the weights initialized randomly.
We randomly initialized the word embeddings to represent the tokens in questions and output action sequences.
We updated the word embeddings within the process of training the Vanilla model and fixed them when training our MARL model.
Therefore the primary and the candidate questions in the training dataset were represented as the sum of the vector for each token.
The DSSM model took such representation of the questions as the input to compute the semantic similarity between the questions. 

In the programmer learning stage, we employed Reptile~\cite{nichol2018reptile} for fast and simple implementation of MAML while avoiding the significant expense of computing second-order derivatives. 
We set $\eta_1 = 1e$-4 when adapting the model to each new task, and set $\eta_2 = 0.1$ to optimize $\vtheta$ with the gradient update derived from the meta-test data.
The reward that the adaptive programmer gained was used to update the retriever parameter $\vphi$ through the AdaBound optimizer~\cite{DBLP:conf/iclr/LuoXLS19} in which the learning rate $\eta_3$ was initially set to $1e\text{-}3$ and the final (SGD) learning rate was set to $0.1$. 

When finding the top-$N$ support set, we set $N = 5$.
For each question, we generated five action sequences to output the answers and rewards.
Adam optimizer was applied in RL to maximize the expected reward.

In the retriever learning stage, we employed the REINFORCE model to optimize the non-differentiable objective directly.
The baseline $\mathcal{C}$ used in REINFORCE was set with a constant value of $1e\text{-}3$ to reduce the variance.
We sampled $M=5$ different sets of the secondary questions and got one unique adaptive programmer for each set with the same meta-learning configuration in the programmer learning stage.
As solving the entity linking problem is beyond the scope of this work, we separately trained an entity/class/relation linker, achieving an accuracy of 99.97\%, 91.93\%, and 94.29\%, respectively. 
When training the MARL model, the predicted entity/class/relation annotations, along with natural language questions, were used as the input sequence.
We trained the MARL model with the batch size of 1 and stopped training when the accuracy on the validation set converged (at around 30 epochs).
We release the source code at \url{https://github.com/DevinJake/MARL} to facilitate replication.

\subsection{Performance Evaluation}

We compared our model with two baseline methods on the CQA task: KVmem~\cite{saha2018complex} and CIPITR~\cite{saha2019complex}. 
Saha et al.~\shortcite{saha2018complex} propose KVmem, a baseline CQA model that combines Hierarchical Recurrent Encoder-Decoder (HRED) with a Key-Value memory (KVmem) network.
The KVmem model retrieves the most related memory to predict the answer from candidate words by attending on the encoded vectors stored in the memory.

CIPITR~\cite{saha2019complex} takes a further step that employs the NPI approach to solving the CQA task without annotations.
CIPITR designs high-level constraints to guide the programmer to produce semantically plausible programs for a question. 
It is worth noting that CIPITR separately trains \emph{a separate model for each of the seven question categories}, and selects the corresponding model to answer questions of the relevant type.
We denote the model learned in this way as \textbf{CIPITR-Sep}.
Besides, CIPITR also trains \emph{one single model over all types of training examples} and uses this single model to answer all questions.
We denote this single model as \textbf{CIPITR-All}.
 

Also, we compared our full model, MARL, with several model variants to understand the effect of our retriever and meta-learner. 
Specifically, \textbf{Vanilla} is a BiLSTM-based model further optimized with reinforcement learning. 
Both Random and Jaccard are MAML-based models with different retrievers. 
\textbf{Random} denotes the model with a retriever that randomly selects questions within the same category. 
\textbf{Jaccard} means the model with a non-learning retriever that makes use of Jaccard similarity on question words. 

We ran the open-source code of KVmem and CIPITR to train the model and presented the best result we got.
KVmem does not have any beam search, and both CIPITR and our model employ beam search for predicting the action sequences.
When inferring the testing samples, we used the top beam~\cite{saha2019complex}, i.e., the predicted program with the highest probability in the beam to yield the answer.  

Table~\ref{tab:comparison} below summarizes the results. 
We can make a number of important observations. 

\begin{enumerate}
	\item Our full model MARL achieves the best overall performance of 66.96\% and 77.71\% for macro and micro F1, respectively, outperforming all the baseline models KVmem, CIPITR-All, and CIPITR-Sep. The performance advantage on macro F1 over the three baselines is especially pronounced (47.51, 47.14, and 19.26 percentage points over KVmem, CIPITR-All, and CIPITR-Sep respectively). This is mainly due to the severe imbalance of the CQA dataset. As Table~\ref{tab:comparison} shows, almost 49\% (462K/944K) of the CQA training set belongs to the Simple Question category, while four other categories only account for 10\% or less each. Given such a large distributional bias, KVmem and CIPITR are unable to learn task-specific knowledge adequately. 
	
	\item MARL achieves the best or second-best performance in all the seven categories. Of the three hardest categories (middle part of Table~\ref{tab:comparison}), MARL delivers the best performance in all three types. This validates the effectiveness of our meta-learning-based approach in effectively learning task-specific knowledge. Note that the two categories that MARL performs the best, Comparative Reasoning and Comparative (Count), both account for less than 5\% of the training set, which further demonstrates MARL's excellent adaptability.
	
	\item CIPITR-Sep achieves the best performance in two \emph{easy} categories, including the largest type, Simple Question. For the three hard categories, it performs poorly compared to other models. This further demonstrates the limitation of coarse-grained adaptation. 
	
	\item CIPITR-All, the model that trains over all types of the questions, performs much worse in all the categories than CIPITR-Sep, which learns a different model separately for each question category. For CIPITR-Sep, the results reported for each category are obtained from the models explicitly tuned for that category. 
	A possible reason for CIPITR-All's significant performance degradation is that it is hard for such a one-size-fits-all model to find the weights that fit the training data when the examples vary widely.
	Besides, the imbalanced classes of questions also deteriorate the performance of the model. 
	Different from CIPITR, our model is designed to adapt appropriately to various categories of questions with one model, thus only needs to be trained once. 
	
	\item In general, our MAML-based model variants, Random, Jaccard, and MARL, outperform their non-MAML counterpart, Vanilla, which demonstrates the advantage of our method in learning task-specific knowledge. 
\end{enumerate}

We also conducted an ablation study to examine the effectiveness of the retriever. 
Table~\ref{tab3} presents the result of the ablation study. 
As can be seen, the random retriever slightly improves performance over the base model Vanilla by 0.97 percentage points, and the fixed retriever Jaccard improves upon Vanilla by 1.59 percentage points. 
Our full model achieves a 2.99 percentage point improvement over Vanilla.
We can observe that the full model improves the overall micro F1 score by 1.40 percentage points compared with the fixed retriever Jaccard.
The performance disparity between the full model and the Jaccard retriever can be attributed to our joint training strategy, in which our full model alternately optimizes the programmer and the retriever. On the contrary, less optimally, the Jaccard retriever is separately trained before training the programmer. 
These results demonstrate the effectiveness of our meta-learning approach as well as the power of our model that jointly optimizes the retriever with the programmer.

\section{Related Work}
\label{Related Work}
%
\textbf{CQA.} A behavior cloning based method, Dialog-to-Action (D2A)~\cite{guo2018dialog}, is proposed to answer complex questions by learning from the annotated programs.
D2A employs a BFS algorithm to annotate the questions with the corresponding action sequences, and use the annotations to train the programmer.
%
%
%
On the other hand, the NPI based methods, i.e., Neural-Symbolic Machines (NSM)~\cite{liang2017neural}, CIPITR~\cite{saha2019complex}, and Stable Sparse Reward based Programmer (SSRP)\cite{ansari2019neural}, employ the yielded answers as the distant-supervision to learn a programmer.
NSM is proposed to answer the multi-hop questions.
It first annotates the questions with the pseudo-gold programs and then assigns the annotated programs with a deterministic probability, therefore, to anchor the model to the high-reward programs.
CIPITR and SSRP both aim to alleviate the sparse reward problem that appears in conventional NPI approaches and employ high-level constraints to guide the programmer to produce semantically plausible programs.
Except for CIPITR, the NPI approaches learn a one-size-fits-all model for the entire dataset.
However, CIPITR learns a one-size-fits-all model for each question type instead of empowering the model to learn to transfer from one question type to another.
%
Different from learning a one-size-fits-all model, we aim to learn a model that quickly discovers the knowledge specific to a new task, and employ the acquired knowledge to adapt the programmer to the new task.

\textbf{Meta-Learning.} The Meta-learning approaches utilize the inductive biases, which are meta-learned in learning similar tasks to make the model learn the new task quickly.
To make the model sensitive to the new task, one popular direction of meta-learning is to train a meta-learner to learn how to update the parameters of the underlying model~\cite{DBLP:journals/corr/LiM17b,DBLP:conf/iclr/HaDL17}, which has been investigated in MAML~\cite{finn2017model}.
In semantic parsing tasks, Huang et al~\shortcite{huang2018natural} 
propose a relevance function to find similar samples to form a pseudo-task for each WikiSQL question.
Subsequently, they reduce a supervised learning problem into a meta-learning problem and employ MAML to adapt the programmer to each pseudo-task. 
Likewise, S2A~\cite{DBLP:conf/acl/GuoTDZY19} separately trains the retriever and the programmer by using the pseudo-gold annotations. 
Based on the similar samples found by the retriever, S2A establishes a meta-learning task for each question and thus employs MAML to finetune the programmer.
Unlike them, we propose a MAML-based approach that trains the retriever and the programmer jointly.
It is worth noting that S2A aims to answer conversational questions while we consider answering single-turn questions. 
Therefore we do not include S2A as a baseline method in the evaluation as it is not directly comparable to our problem setup.

\section{Conclusion}
\label{sec:conclusion}
In this paper, we presented a novel method for complex question answering over knowledge bases. 
In a meta-learning framework, our model jointly and alternately optimized a retriever, which learned to select questions, and a programmer, which learned to adapt to the selected secondary questions to produce an answer to a given primary question.
Our model was capable of quickly adapting to new questions as it could learn from similar questions. 
Moreover, it did so from weak supervision signals, the model's performance on question answering. 
Thus, our model addressed several essential challenges facing existing methods, namely the significant distributional biases present in the dataset and the high cost associated with manual labeling of similar questions. 
Our evaluation against a number of state-of-the-art models showed the superiority of our model on the large-scale complex question answering dataset CQA.
In the future, we plan to extend our model to other domains and tasks that require the manual construction of support sets.

\section*{Acknowledgments}
Research presented in this paper was partially supported by the National Key Research and Development Program of China under grants (2017YFB1002801, 2018YFC0830200), the Natural Science Foundation of China grants (U1736204, 61602259), Australian Research Council (DP190100006), the Judicial Big Data Research Centre, School of Law at Southeast University, and the project no. 31511120201 and 31510040201.

\clearpage
\bibliographystyle{named}
\bibliography{ijcai20}
\end{document}